\documentclass[sigconf]{acmart}
\usepackage{times}
\usepackage{soul}
\usepackage{url}
\usepackage[utf8]{inputenc}
\usepackage{caption}
\usepackage{tikz}
\usepackage{amsmath,amsfonts, bm}
\usepackage{algorithm}
\usepackage{algorithmicx}
\usepackage[noend]{algpseudocode}
\usepackage{mdframed}

\usepackage{graphicx}
\usepackage{textcomp}
\usepackage{booktabs} 
\usepackage{siunitx}
\usepackage{xspace,url}

\usepackage{color}
\usepackage{colortbl}
\usepackage{multirow}
\usepackage{multicol}
\usepackage{makecell}
\usepackage{tabularx}
\usepackage{listings}
\usepackage{graphicx}
\usepackage{caption}
\usepackage{enumitem}
\usepackage{adjustbox}
\usepackage{subcaption}
\usepackage{listings}
\usepackage{xcolor}

\usepackage[colorinlistoftodos]{todonotes}

\acmSubmissionID{\#128}

\definecolor{airforceblue}{rgb}{0.36, 0.54, 0.66}
\newcommand{\wh}[1]{\textcolor{black}{#1}}    

\newcommand{\name}{AutoDroid-V2\xspace} 

\newcommand{\ie}{\textit{i}.\textit{e}.~}
\newcommand{\eg}{\textit{e}.\textit{g}.~}

\author{Hao Wen$^{1}$, Shizuo Tian$^{1}$, Borislav Pavlov$^{1}$, Wenjie Du$^{1,*}$, Yixuan Li$^{1,*}$, Ge Chang$^{1,*}$, Shanhui Zhao$^{1,*}$, Jiacheng Liu$^{1,*}$, Yunxin Liu$^{1,2}$, Ya-Qin Zhang$^{1}$, Yuanchun Li$^{1,2,3,\dagger}$}
\thanks{$\dagger$ Corresponding author: Yuanchun Li (liyuanchun@air.tsinghua.edu.cn).}
\thanks{* Work done during internships at AIR, Tsinghua University.}
\affiliation{$^1$~Institute for AI Industry Research (AIR), Tsinghua University \\ \:$^2$~Shanghai Artificial Intelligence Laboratory\\ \:$^3$~Beijing Academy of Artificial Intelligence (BAAI) 
\country{}}

\renewcommand{\shortauthors}{Wen et al.}

\begin{document}

\acmYear{2025}\copyrightyear{2025}
\acmConference[MobiSys '25]{The 23rd Annual International Conference on Mobile Systems, Applications and Services}{June 23--27, 2025}{Anaheim, CA, USA}
\acmBooktitle{The 23rd Annual International Conference on Mobile Systems, Applications and Services (MobiSys '25), June 23--27, 2025, Anaheim, CA, USA}
\acmDOI{10.1145/3711875.3729134}
\acmISBN{979-8-4007-1453-5/25/06}


\title{AutoDroid-V2: Boosting SLM-based GUI Agents via Code Generation}




\begin{abstract}
Large language models (LLMs) have brought exciting new advances to mobile UI agents, a long-standing research field that aims to complete arbitrary natural language tasks through mobile UI interactions. 
However, existing UI agents usually demand powerful large language models that are difficult to be deployed locally on end-users' devices, raising huge concerns about user privacy and centralized serving cost. 
Inspired by the remarkable coding abilities of recent small language models (SLMs), we propose to convert the UI task automation problem to a code generation problem, which can be effectively solved by an on-device SLM and efficiently executed with an on-device code interpreter. 
Unlike normal coding tasks that can be extensively pre-trained with public datasets, generating UI automation code is challenging due to the diversity, complexity, and variability of target apps. 
Therefore, we adopt a document-centered approach that automatically builds fine-grained API documentation for each app and generates diverse task samples based on this documentation. 
By guiding the agent with the synthetic documents and task samples, it learns to generate precise and efficient scripts to complete unseen tasks. Based on detailed comparisons with state-of-the-art mobile UI agents, our approach effectively improves the mobile task automation with significantly higher success rates and lower latency/token consumption. Code is open-sourced at \url{https://github.com/MobileLLM/AutoDroid-V2}. 

\end{abstract}

\ccsdesc[500]{Human-centered computing~Ubiquitous and mobile computing}
\ccsdesc[300]{Computing methodologies~Artificial intelligence}

\keywords{Mobile UI Agent, Small Language Models, Mobile Task Automation, Script Generation}

\maketitle

\section{Introduction}

Automating the control of mobile devices through natural language has long been a goal for mobile developers and researchers \cite{Put-that-there, ulink, li2024personal}. Agents powered by Large Language Models (LLM) and Vision Language Models (VLM, both LLM and VLM are refered as LLM for simplicity in this paper) have emerged as promising solutions for automating flexible and complex user tasks~\cite{foundatioin_model_as_firmware, autodroid, you2024ferret, mobileagent, lee2023explore, liu2024autoglm, ui_agent_survey}. 
These agents function by interpreting user instructions, analyzing the graphical user interface (GUI) states, and predicting UI actions (e.g., touch, scroll, input text, etc.) to be performed in the target apps iteratively. 
Most of these LLM-based GUI agents operate in a step-by-step manner, \ie querying the LLM at each GUI state for dynamic decision and reflection. They typically include the user's task and the observed GUI state (e.g., a screenshot or a text-based view hierarchy) in the LLM prompt, requesting the model to generate the required UI action. This iterative process continues until the task is completed. For this reason, we refer to these conventional designs as \textbf{``Step-wise GUI agents''} in this paper.

While step-wise GUI agents hold great potential, they mostly rely on powerful reasoning and reflection abilities, which are dominated by large cloud-based models (\eg GPT-4o, Claude, etc.). Due to the high computational cost and/or closed-source nature of these models, they can hardly be deployed to mobile devices. This situation can lead to significant privacy and security risks of sharing each personal GUI page to the cloud~\cite{privacy, ma2024caution, zhang2024attacking}, high user-side traffic consumption, and high server-side centralized serving costs, hindering large-scale deployment of GUI agents.

Among the possible countermeasures to this problem, \textbf{small language models (SLMs)} (There is no clear boundary between LLM and SLM. In this paper, we use SLM to represent the LLMs that can be deployed on the device.) and \textbf{efficient inference systems} have emerged as promising directions to enable on-device deployment of more powerful models~\cite{slm_survey, ondevice_survey, slm_survey_xu}. The current capabilities of these SLMs are still far from their cloud-based counterparts, and the capability gap is anticipated to remain in the near future due to the scaling laws~\cite{scaling_law}.
\textbf{Customizing a domain-specific SLM} can also be a potential solution. However, training a domain-specific model requires a large-scale, high-quality labeled dataset covering diverse usage scenarios, which are difficult to obtain in the UI task automation domain. Most existing mobile task automation datasets~\cite{motif, llamatouch, rawles2024androidworld, metagui} only have hundreds of human demonstrations. The largest dataset AitW~\cite{AITW_dataset} has 715k crowd-sourced task demonstrations, while the quality is low for most demonstrations (we compare the agents trained with such datasets in Section~\ref{sec:accuracy}). Meanwhile, the LLM abilities used in existing GUI agents are diverse, including not only simple GUI grounding abilities that can be offered by a small tailored model ~\cite{cheng2024seeclick, cogagent, gui_grounding}, but also general reasoning and reflection abilities that can hardly be distilled to a much smaller model~\cite{ui_agent_survey, mobileagentv2, bishop2024latent}.


Here, we argue that the widely adopted step-wise mechanism of GUI agents may need to be changed in order to accommodate the requirements of practical on-device agents.
First, step-wise decision-making requires high and frequent reasoning and reflection about the current state and progress, which are essentially challenging for SLMs.
Second, even if the step-wise decisions can be made with reasonable accuracy, calling the model at each step is very compute-intensive, making it difficult to provide a smooth user experience with limited on-device computational resources.

Fortunately, we observe that \textbf{SLMs can be trained to have a unique but important strength in coding}, \ie producing executable code based on user instructions~\cite{phi1, codellama, zhu2024deepseek-coder}. This strength is largely attributed to extensive training on code datasets~\cite{llama3, mistral2}, as it is relatively easier to collect large-scale, high-quality source code data compared to many other types of data. Remarkably, some on-device SLMs now achieve accuracy levels comparable to closed-source LLMs~\cite{zhu2024deepseek-coder}. This raises an interesting question: \emph{could the step-wise decision-making problem in GUI task automation be reframed as a code generation problem to better leverage the capability of on-device SLMs?}

\begin{figure}
  \centering
  \includegraphics[width=1.0\linewidth]{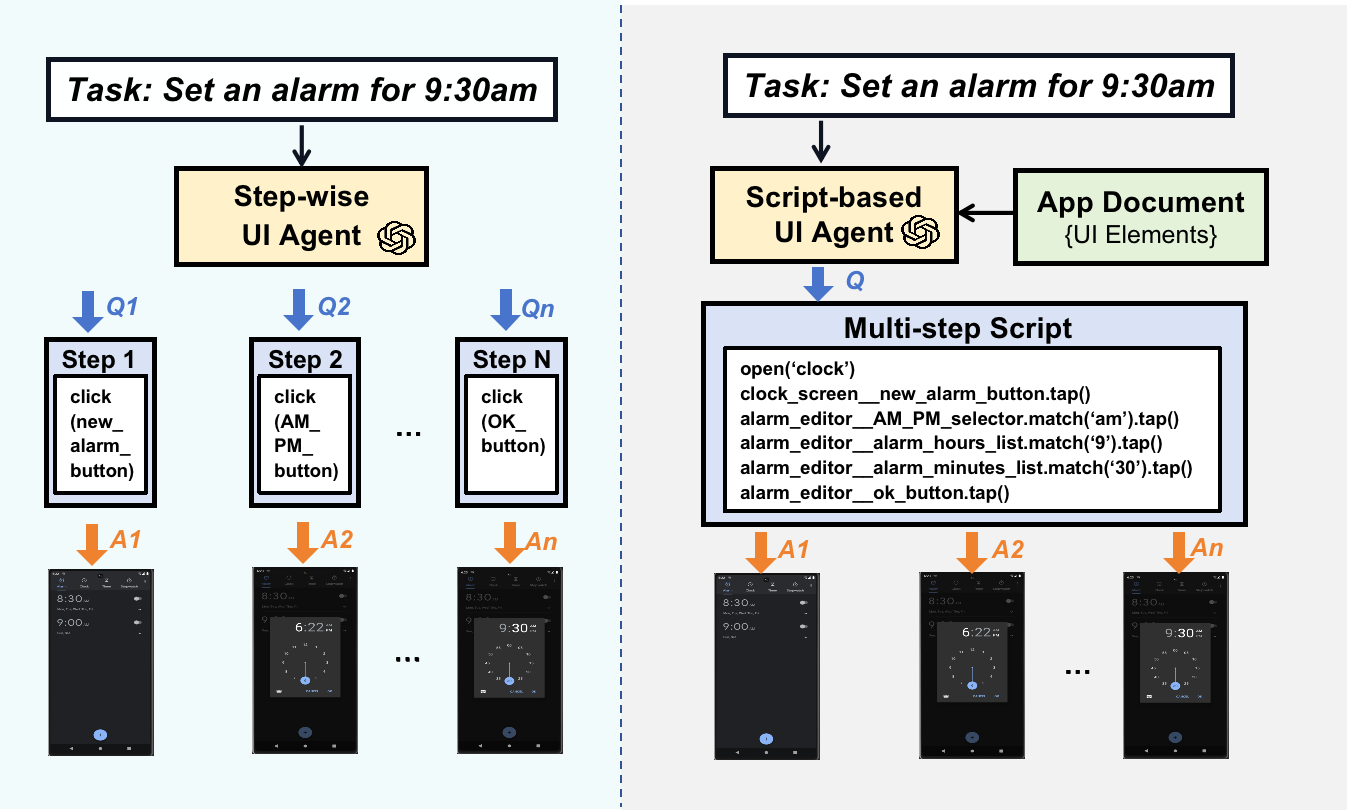}
  \caption{Comparison of conventional step-wise GUI agent and our script-based GUI agent. \textnormal{The step-wise GUI agent includes the user’s task and the current GUI state in the LLM prompt ($Q1, \dots Qn$), asking the model to generate a GUI action for each step ($A1, \dots An$). The script-based agent inputs the task and app document ($Q$), prompting the LLM to generate a multi-step script for execution.}}
  \label{fig:differences}
\end{figure}

To this end, we introduce \name to investigate \textbf{how to build a powerful GUI agent upon the coding capabilities of SLMs}. The key difference of \name is to complete GUI tasks by generating and executing multi-step scripts based on user instructions, rather than deciding one action at a time. \wh{We refer to this approach as the \textbf{``Script-based GUI Agent''}, which largely differs from traditional step-wise GUI agents, as illustrated in Figure \ref{fig:differences}. }
Such a script-based agent can effectively address the aforementioned two problems of step-wise approaches: 
(1) {Efficiency}: Agents can generate a single script for a series of GUI actions to complete a task based on the user task, significantly reducing query frequency and consumption.
(2) {Capability}: Script-based GUI agents rely primarily on the coding ability of SLMs, which have been proven effective by numerous existing studies on lightweight coding assistants.

A natural question that may be raised for script-based GUI task automation is why it is feasible to generate accurate scripts before knowing the actual situation at each runtime step. 
Conceptually, each app, although it may be very complicated, can be viewed as a finite state machine, where the number of states is predetermined in the app development stage. Thus, in principle, a human or AI with full information about the state machine is able to complete tasks without observing each state.
Practically, realizing \name requires to further address two challenges.
(1) \textbf{Understanding GUI elements}: It is necessary - but challenging - for LLMs to understand all GUI elements and their functions within an app to generate an accurate script. This difficulty arises from the complexity of mobile apps, which often involve numerous GUI states, each featuring over 100 GUI elements (e.g., buttons, icons, inputs) organized in a hierarchical structure, as well as complex transition relationships between GUI states. 
(2) \textbf{Dynamic Nature of Mobile apps}: Many GUI states and GUI elements can change in text, location, or size (\eg contact names, note titles, dates). As a result, the generated script may become non-executable if a GUI element changes and cannot be located, due to the dynamic nature of GUI elements. 


\name addresses these challenges by automatically generating a fine-grained yet concise app document, which guides agents in generating precise multi-step code to interact with UI elements and complete user tasks. The method consists of two stages: UI-Centric App Documentation and Document-Guided Code Generation. 
The {UI-Centric App Documentation} stage is responsible for analyzing the UI functionalities based on random exploring histories and recording the interactive elements. 
{The {Document-Guided Code Generation} stage automatically generates code to interact with GUI elements, leveraging a domain-specific language (DSL) to efficiently handle runtime dynamicity.}

We evaluate \name in two benchmarks with real execution environments and a total number of 226 tasks across 23 mobile apps, in comparison with strong baselines including AutoDroid \cite{autodroid}, SeeClick \cite{cheng2024seeclick}, CogAgent \cite{cogagent}, and Mind2Web \cite{mind2web}. 
The results demonstrate that \name can effectively enhance the accuracy and efficiency of on-device GUI agents, achieving 10.5\%-51.7\% higher task completion rate than baselines, reducing runtime input and output token consumption by 43.5x and 5.8x, reducing LLM inference latency by 5.7-13.4× compared to baselines. 

Our work makes the following technical contributions:
\begin{enumerate}
    \item \wh{We introduce the first GUI agent driven by document-guided multi-step program generation, 
    presenting a promising solution combining the remarkable coding capabilities of SLMs and efficient code-based task execution. }
    \item We introduce a scalable automated app document generation method that can compress UI information from an app exploring history, as well as an interpreter that ensures reliable and efficient script execution.
    \item We conduct extensive experiments to demonstrate the efficiency and effectiveness of our approach, offering insights into the advantages and limitations of our script-based GUI agents and existing step-wise approaches.
\end{enumerate}

\section{Related Work and Motivation}
\label{section:background}

\subsection{GUI-based Mobile Task Automation}
The goal of mobile task automation is to automatically complete user tasks by interacting with mobile devices. 
\wh{A GUI-based task automation agent takes a natural language task related to a mobile app and outputs a sequence of GUI actions executable on a smartphone. }
A \textbf{task} is a multi-step functionality request from the user intended for completion on a smartphone, often lacking explicit instructions. 
\wh{A \textbf{GUI State} reflects the current status or condition of the user interface in the app, usually structured as a GUI tree of controls represented by images and text. }
A \textbf{GUI element} is a control in the GUI state that human users can interact with, such as a button, a text box, an input field, and a slider. 
A \textbf{GUI action}, performable by the user or an agent on the device screen, is defined by (GUI element, action). Action represents how the target GUI element is manipulated, including ``click'', ``long click'', ``input <text>'', and ``swipe <direction>''. 

\subsection{Related Works}

\wh{\textbf{LLM-based GUI Agents. }}
\wh{With the rise of large foundation models, there has been a trend leveraging LLMs~\cite{llm_brained_ui_agent, wen2023droidbot, autodroid, mobilegpt, responsible_task_automation} and Vision Language Models (VLMs)~\cite{auto-ui, mm-navigator, yang2023appagent} to guide mobile task automation. These models enable more flexible and scalable task automation, typically executing tasks in a step-wise manner. 
Some research efforts focus on code generation for web page task automation~\cite{gur2024a, tao-etal-2024-webwise}, relying primarily on the HTML to generate actions within the current screen. In contrast, \name adopts a full task-level script generation approach for mobile GUI agents.
LLM-based GUI agents often rely on sensitive data—like screenshots and interaction logs—that may be sent to remote servers. This creates serious privacy risks, including data breaches and misuse~\cite{ui_agent_survey}. High data traffic and server costs also hinder large-scale deployment. On-device LLM-powered agents are therefore essential for secure and affordable mobile task automation.}


\wh{\textbf{SLM-based GUI Agents. }}
\wh{Researchers have explored various techniques to customize domain-specific SLMs for GUI-based mobile task automation. One area of focus involves enhancing the grounding capability of SLMs, which refers to locating GUI elements based on user instructions within a single GUI screen~\cite{cheng2024seeclick, cogagent, gui_grounding}. These methods gather extensive GUI-domain training data and use it to fine-tune a vision language model, thereby improving their ability to follow GUI domain-specific instructions. Other studies aim to acquire domain-specific knowledge about particular apps through exploration and use it to enhance SLMs' task-solving capabilities~\cite{autodroid, bai2024digirl}. 
However, the limited reasoning abilities of SLMs and large computational resource consumption hinder their widespread use, and we try to improve the efficiency and effectiveness of these methods by introducing document-guided script generation. }

\subsection{Script-based GUI Agents: Opportunities and Challenges}

The limitations of current methods highlight the need for a more accurate and efficient approach to automating mobile UI tasks beyond traditional step-wise UI agents. Recent studies have shown that SLMs can perform well on challenging coding tasks \cite{zhu2024deepseek-coder,codellama}, and synthesizing executable code is more advantageous than using Text/JSON actions in agents~\cite{wang2024executable}. This motivates the development of a script-based on-device GUI agent to address the identified limitations, offering the following benefits: 

(1) {Reduction of Query Overhead: }Instead of querying the model for each individual step, script-based GUI agents can generate a multi-step action plan based on a single piece of code. This approach significantly reduces the overhead associated with multiple queries. 
(2) {Enhanced Domain-specific Knowledge Utilization: }By building an API document specifically for code generation, GUI agents can better leverage app-domain knowledge in a manner more familiar to language models. 
(3) {Improved Action Tracking and Termination Control: }Executing GUI actions based on code lines shifts the responsibility of action tracking from the model to a code status tracker. This approach is more controllable and avoids the issues of misalignment and hallucination, leading to more reliable task completion.

Although there is great potential for script-based on-device GUI agents, it still faces significant challenges. 

\textbf{Dynamic nature of mobile apps. }
For mobile apps, many UI states and elements are dynamic, frequently changing in text, location, or size (e.g., titles, dates, names). 
This variability increases the difficulty of both extracting app-domain knowledge and executing auto-generated scripts. 
To generate multi-step scripts that can be directly executed within app UIs, the system needs to provide all UI information within the entire app for summarizing knowledge. These dynamic UI elements need to be identified and abstracted to give an overview of the app. When executing the script, the system needs to locate the dynamic UI elements and perform actions according to the code, even when the elements change.

\textbf{Extensive UI element space. }
A typical UI may contain over 100 elements with different XML descriptions, expanding the element space in the exploration history to tens of thousands. 
Describing all these elements for generating code is both redundant and potentially misleading, complicating the task of identifying the necessary UI elements for a given task. 
Additionally, UI elements often have relationships, such as inclusion (where some elements are nested within others, \eg a song's title and artist within a single song item in Figure \ref{fig:build-ui-element}) or dependency (where one UI element is editable only after another has been clicked). 
It is still difficult for LLMs or SLMs to understand complex element relationships only based on a static UI document. 

\textbf{Limited Context Length and Computational Resources}
Due to limited computational power, reduced memory capacity, and energy constraints, on-device SLMs often have shorter context lengths (e.g., 4096 for Llama 3.1 8B~\cite{llama3}). As apps grow more complex, including detailed app usage patterns in the prompt can easily exceed the context length. Without this usage pattern, script generation loses valuable guidance and insight into app usage. 
Moreover, for a given user task, only a small portion of GUI elements may be relevant, yet they can be buried within a long app usage context. In such cases, the on-device SLMs may struggle to identify and utilize the relevant information~\cite{jiang-etal-2024-longllmlingua}.

\section{Our Approach: \name}
\label{sec:approach}
\begin{figure*}
  \centering
  \includegraphics[width=0.95\linewidth]{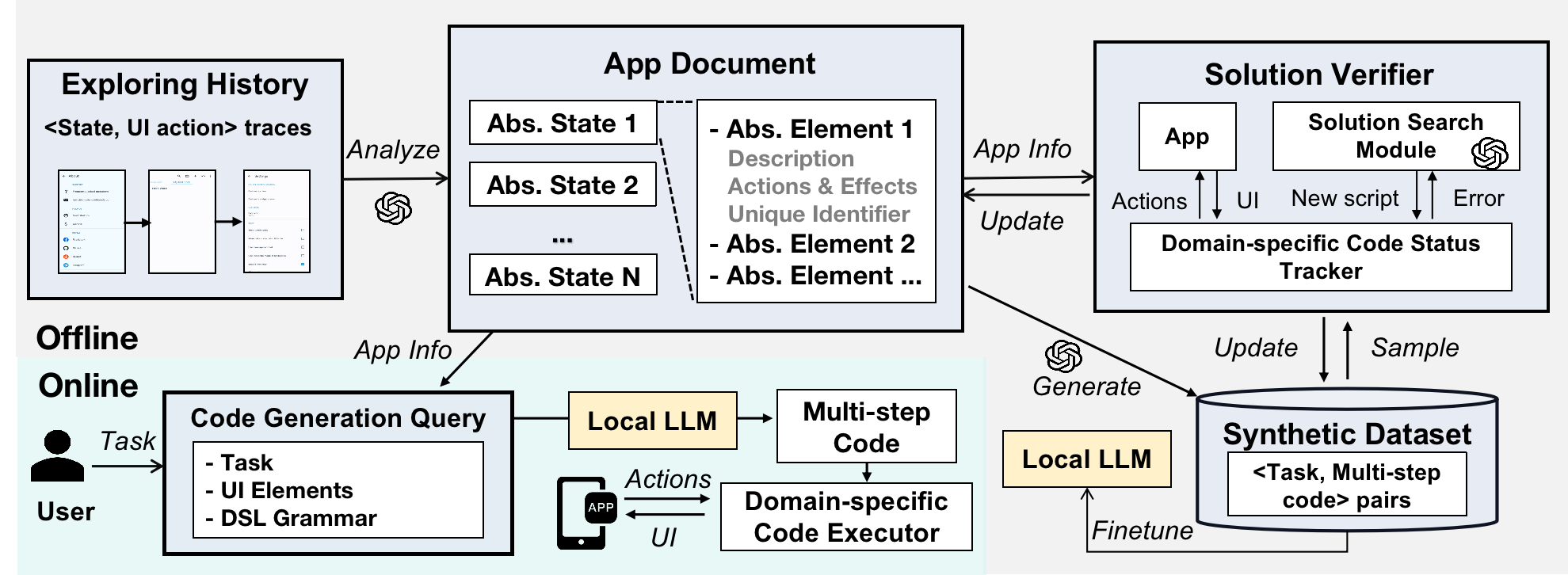}
  \caption{The architecture of \name.}
  \label{fig:main}
\end{figure*}

We introduce \name to address the aforementioned challenges of on-device GUI agents. The core idea behind \name is automatically constructing an informative app document and guiding the LLMs/SLMs to generate domain-specific device-control scripts.

As shown in Figure \ref{fig:main}, \name operates in two stages. 
\textbf{Offline Stage:} \name first constructs an app document by analyzing the app exploration history (\emph{details in \S\ref{sec:doc_gen}}). The document serves as the fundamental guidelines for flexible script generation, which is desired to be concise (easy to process by LLMs) and precise (helpful for generating complicated executable code). We introduce various techniques (AI-guided GUI state compression, element XPath auto-generation, GUI dependency analysis, etc.) to achieve these goals.
The built document is then used to synthesize large-scale user tasks for fine-tuning the local LLM (\emph{details in \S\ref{sec:tune_llm}}). Given the importance of training data in LLM customization, we adopt sandbox validation and tree-based search techniques to improve the data quality.
\textbf{Online Stage:} Given a user's task request, \name calls the customized local LLM to generate a multi-step script and executes the script with our domain-specific interpreter (\emph{details in \S\ref{sec:runtime}}). Running the script reliably (\ie able to handle runtime dynamics) and efficiently are two main objectives of the interpreter, which are addressed with our dependency-enhanced execution and prompt compression/caching techniques.

\subsection{Automated App Document Generation}
\label{sec:doc_gen}

\begin{figure*}
  \centering
  \includegraphics[width=0.9\textwidth]{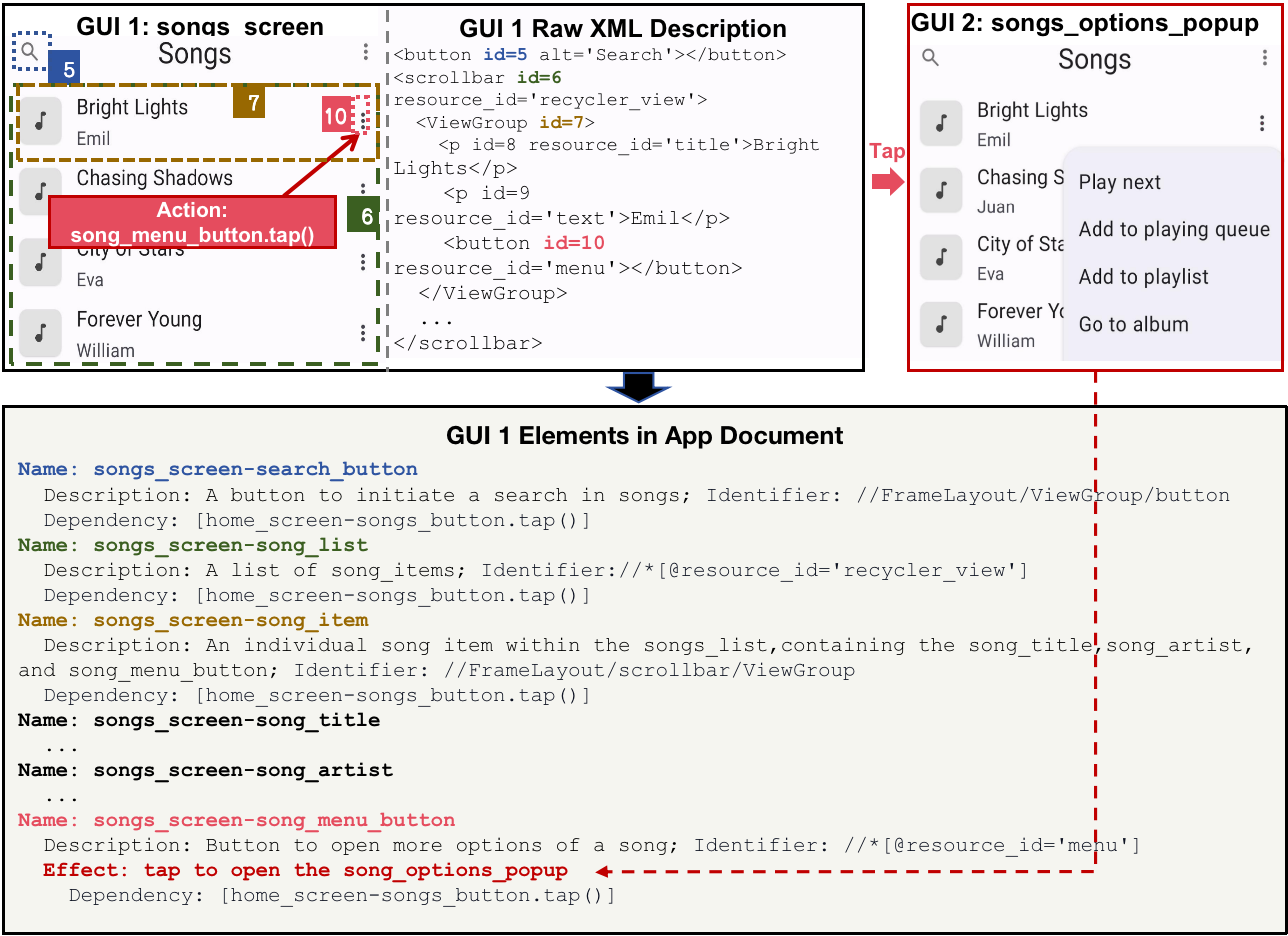}
  \caption{The document of an app containing essential abstract elements in one GUI. }  
  \label{fig:build-ui-element}
\end{figure*}

The goal of automated app document generation is to capture and analyze essential GUI elements in an app, guiding the backbone SLM to be aware of possible interaction use cases even without real-time observation. The input consists of random exploration traces, modeled as sequences of <GUI State, GUI Action> pairs. 
\wh{Following existing works~\cite{autodroid, yang2023appagent, mobilegpt}, we suppose that the exploration traces can be generated by GUI testing tools~\cite{droidbot} that explore apps randomly or through stochastic greedy depth-first strategies, or by a user trace monitor that records app usage in the background~\cite{create_accessibility_service}. The exploration phase operates independently of user instructions, and the resulting traces are often redundant or aimless, containing many repeated actions and GUI elements.}
The output is a concise app document containing typical GUI states, important elements, and GUI transition relationships, as shown in Figure~\ref{fig:build-ui-element}. Two primary challenges arise in this process: reducing the GUI element space and modeling transition relationships between GUI elements.

\subsubsection{Functionality-aware GUI States Grouping} 
To ensure a structured and meaningful representation of the app's interface and interaction logic, \name first organizes the GUI states by iterating over the exploration history. Specifically, \name merges similar GUIs into an abstract GUI state, where similarity is determined by both \textbf{layout similarity} and \textbf{functionality similarity}. 

Layout similarity refers to how the visual components (e.g., buttons, text fields, etc.) are arranged within the GUI. Let ${T_1, T_2, ... T_n}$ represent a set of GUI trees observed in the exploration traces. The layout of each tree, $T_i$, is extracted by removing detailed content and repeated sibling elements. If a group ${T_1, T_2, ... T_n}$ shares the same layout $l_i$, they are considered layout-similar. 

However, layout similarity alone is insufficient. Many different GUI states in an app may share the same layout due to common development patterns. For instance, a date selection screen and a background theme selection screen in a calendar app might both use a similar layout (e.g., a popup with a series of checkboxes). Despite having the same layout, these two screens serve different functions. 
To address this, we introduce functionality similarity, which considers the tasks or actions that users can perform within a particular GUI state. This ensures that even if the layouts are similar, the underlying functionality is accurately captured and differentiated. 
\name uses LLMs (\eg GPT-4o) to assess and classify each GUI group based on its functionality after the layout-based grouping is performed. 
\wh{Specifically, for each step in the exploration history, \name constructs a prompt that includes the functionalities of previous GUI states (initially an empty list), and the current GUI state. It sends the prompt to the LLM to determine whether the current GUI state corresponds to any of the previously identified functionalities. If a match is found, the LLM maps the current GUI state to that existing functionality. If not, it generates a new name and description, which are then added to the functionality list. } 

% ...






Then, we can represent each abstract GUI state $S^{(i)}$ after grouping as $(functionality^{(i)}, layout^{(i)})$, where $functionality^{(i)}$ refers to the task or function associated with the GUI state (e.g., "date selection" or "theme selection") and $layout^{(i)}$ represents the shared visual structure.

\subsubsection{Abstract GUI Elements}
To reduce the GUI element space, we introduce abstract GUI elements in \name, each of which may encapsulate several real GUI elements encountered during exploration. 
A \textbf{single abstract element} is typically one visible node in the GUI tree with which the user can interact. These elements can be classified into two types: \textbf{dynamic} and \textbf{static} elements. Static elements are those that remain constant regardless of changes in the app's status, such as the `Search button' in Figure \ref{fig:build-ui-element}. Dynamic elements contain content that can change based on context, such as the `song title' in Figure \ref{fig:build-ui-element}. 

In addition to a single abstract element, \name introduces another abstract GUI element type called the \textbf{abstract element list}. This represents a collection of single elements that can be indexed or filtered, such as the `song list' in Figure \ref{fig:build-ui-element}. 

To address the dynamic nature of mobile apps and further reduce the GUI element space, \name merges a group of dynamic GUI elements (such as a group of specific songs in Figure~\ref{fig:build-ui-element}) into one abstract GUI element (song\_item), designating the parent node of these dynamic elements as the element list (song\_list). We believe element list type can represent a large set of single elements in the app GUI because agents can interact with any of it by giving commands such as \textit{<element\_list>.match(`name')} or \textit{<element\_list>[<idx>]} as mentioned in Section~\ref{sec:dsl}. This abstraction allows us to represent a large set of elements without needing to include every detailed element\_list item in the document.

\begin{table}
	\caption{Components of Abstract GUI Elements}
	\centering
	\resizebox{.48\textwidth}{!}{
	\begin{tabular}{p{1.8cm}p{5.6cm}p{2.8cm}}
		\hline
		\textbf{Component} & \textbf{Description} & \textbf{Example} \\
		\hline
		Name & A concise name for the element formatted as \textit{state\_name-element\_name}. & songs\_ui-song\_list \\
		\hline
		Description & A brief description of the element's purpose and functionality. & A list of song\_items \\
		\hline
		Identifier & Uniquely identifies and locates a specific GUI element within the app's GUI tree. & //*[@resource id=`recycler view'] \\
		\hline
		Options & All the possible keywords of the items in this element list. The items can be chosen by using `<element\_list>.match(Options[i])' statement, which will be explained in Section \ref{sec:dsl}. & <song\_names> \\
		\hline
        Effect & describes the effect of performing actions on it. & tap to open the song\_options\_popup \\
            \hline
		Dependency & The navigation path for one GUI element from the other GUI screens. & Tap main\_screen-open\_song\\
		\hline
        
	\end{tabular}
	}
    \label{tab:element_properties}
\end{table}

As shown in Table~\ref{tab:element_properties}, an abstract element comprises four necessary components: \textit{Name, Description, Identifier, Dependency} along with two optional components: \textit{Action Effect, Options}. 
For the \textbf{element name}, the GUI state name is used as a prefix because there may be several elements with the same name that belong to different states and function differently. For example, tapping the 'more options' button in different GUIs can lead to different GUI states. \textbf{Identifier} is used during runtime to locate and interact with the desired GUI element based on the rules defined in the script. \name uses XPath (XML Path Language) as the identifier for each GUI element. Due to the frequent changes in the number and properties of GUI elements, using a single XPath to locate the GUI element can be unreliable at runtime. 
\wh{Multiple XPaths form a prioritized queue for identifying GUI elements. The first (most precise) XPath specifies all element properties (\ie $t_e$ (text), $alt_e$ (alternate text), $r_e$ (resource id of e), $path_e$ (relative position of a GUI element from the root of the GUI tree)\}, but may fail if some properties change at runtime. Subsequent XPaths omit changeable properties$\{alt_e, r_e, path_e\}, \{r_e, path_e\}$, etc. ($path_e$ is not used alone in the XPath because of the repeatness of UI elements). During execution, the system attempts to ground the GUI element using the XPath queue, starting with the most precise XPath and moving down the queue until a match is found. }

To identify abstract GUI elements through an exploration history, we enumerate all GUI elements within every GUI state and use LLMs to determine the element: \textit{Name, Description, and Options} for important elements. 
The \textit{Effect} of each element is determined by the subsequent GUI state triggered by interaction. Additionally, LLMs are used to categorize elements as dynamic or static and encapsulate them for further processing. 
\wh{Specifically, for each GUI state from one abstract GUI state, \name constructs a prompt (shortened for clarity) to query LLMs (\eg GPT-4o) to classify the GUI elements and their properties. }
\begin{lstlisting}[language=, caption={Structure of GUI element classification prompt.}][htbp]
Imagine you are a UI annotator tasked with describing interactive elements of a mobile application in the form of a UI document. You are given: 
<Current UI State>
<UI Action Performed on Current UI State> (including the functionality of the next GUI state)
<Previously Classified UI Elements in the Same Abstract GUI State>
<An example>
Now, you should output all new UI elements (i.e., those not present in the previously classified elements) that can be interacted with in the current UI state. 
\end{lstlisting}

\subsubsection{Forward and Backward Dependency}

To provide necessary information for agents to plan solution steps, \name profiles the GUI element transition relationships by constructing an Element Transition Graph (ETG) and analyzing both the forward and backward dependency for each element. In the ETG, transitions are represented as triples $(e_i, a_i, e_{i+1})$. Here, $e_i$ and $e_{i+1}$ are nodes representing GUI elements, and $a_i$ is the directed edge representing the action taken to transition from $e_i$ to $e_{i+1}$.To build ETG, we define the root node $e_r$ as the elements in the first GUI of the app after it is opened. For each action $a_i$ observed during the exploration, we locate the existing element $e_i$ that $a_i$ is performed on, and identify all the GUI elements ($e_{i+1}, e_{i+2}...$) that become available in the subsequent GUI state resulting from $a_i$. Edges are added to represent the transitions: $(e_i, a_i, e_{i+1}), (e_i, a_i, e_{i+2}), etc.$ 

After constructing ETG, we analyze the transition relationships between elements by defining two types of dependencies for each element $e_i$: the {backward dependency} and the {forward dependency}. \textbf{Backward dependency} records all the paths from other elements $e_1, e_2, ...e_{i-1}$ leading to $e_i$. \textbf{Forward dependency} represents the subsequent GUI state $e_i$ leads to. At runtime, forward dependency is included in the prompt, providing the context by showing how one GUI element leads to the next sequence of interactions. If the script generated by the LLM fails to correctly trigger a transition between GUI elements, the backward dependency is then used to trace back through the possible previous interactions and identify an alternative path that can successfully reach the target GUI element.


\subsection{Data Synthesis for SLM Customization}
\label{sec:tune_llm}

To improve the task automation capabilities of on-device SLMs, it is crucial to train them with large-scale data. 
We generate a large set of simulated user tasks of varying complexity based on the app document, along with their corresponding solutions. 
These synthetic task-solution pairs are used to fine-tune the SLMs, enabling them to create GUI-specific code with correct syntax and dependencies. 

\subsubsection{Large-scale Task \& Solution Generation}
\label{sec:dsl}

\name generates tasks by sampling groups of elements from the app document, creating task-solution data of varying complexity levels. This approach enables on-device SLMs to learn to handle diverse tasks effectively. Task generation can be viewed as the inverse of task automation. In task automation, a user task $\mathcal{T}$ is mapped to a set of GUI elements and corresponding actions, forming a sub-graph $G_s$ in the Element Transition Graph (ETG). Conversely, generating tasks is to map $G_s$ to $\mathcal{T}$. 
\name iteratively samples a $G_s$ from the ETG and queries LLMs to generate user tasks solvable with these elements. 
The complexity of a task is determined by the number and type of GUI elements involved and the number of actions required to complete the task. 
By varying the size of $G_s$, tasks of different complexities can be generated, denoted as $\{\mathcal{T}_1, \mathcal{T}_2,..., \mathcal{T}_n\}$. 



To regulate the task solution format and execute them on real devices, we develop a Python library to interpret the scripts. 
The design principle of this library is being concise and intuitive, allowing LLMs/SLMs to effectively learn through simple instructions and in-context examples. 
The library provides two categories of functions: GUI action APIs and information retrieval APIs.
The GUI action APIs (including \textit{tap, long\_tap, set\_text, scroll}) are designed to manipulate an element or a specific child element within an element list. These actions are invoked using methods like: \textit{<element>.tap(), <element>.tap(child\_element),  <element>.set\_text(<text>)}, or \textit{<element>.scroll(<direction>)}
Information retrieving APIs (including \textit{get\_text, get\_attributes, match, indexing}) facilitate the accurate identification of task-related GUI elements by gathering necessary information about them. Examples include \textit{<element\_list>.match(<text or attribute dict>)} or \textit{<element\_list>[<idx>]}. 

Then, we design a detailed prompt to guide LLMs in generating executable code of user tasks for training.
The prompt includes the user task, a simplified App Document (which details the name, description, options, and effects of each element), a description of the domain-specific library, and the GUI elements in the current GUI (often the GUI elements in the original GUI screen of the app):  

\begin{lstlisting}[language=, caption={Structure of the code generation prompt.}][htbp]
Imagine that you are an agent using a smartphone to complete a task. You need to write scripts to manipulate the GUI elements in the apps. 

In the script, except for the common Python control flow (for, if-else, function def/calls, etc.), you can use the following APIs: <APIs in GUI domain-specific library>

You can use the following important GUI elements: 
...
<element i>: <description of element i>
Effect: <effect of element i>
Options: <options of element i>
...

The available GUI elements in the current GUI: 
<current available GUI elements>

Your task is: <the user task>

Your answer should follow this JSON format:

{
    "plan": "<a high-level plan to complete the task>",
    "elements": "<analyze the elements that could be used to complete the task>", 
    "script": "<the script to complete the task>"
}

\end{lstlisting}

From the LLM's response, we can then parse an executable script. For example, for the task about the app shown in Figure \ref{fig:build-ui-element}, LLM can generate the following script: 

\begin{lstlisting}[language=Python, caption=LLM generated Python code for the user task. ][htbp]
'''
User task: Add the songs of Artist Emil to the Playlist 'Emil'. 
'''
# Navigate to the Songs section
home_screen-songs_button.tap()
# Flag to control the scrolling loopdone = False
while not done:
  for song_item in songs_screen-song_list: 
    song_artist = song_item.get_text(songs_screen-song_artist)
      if song_artist == 'Emil':
        song_item.tap(songs_screen-song_menu_button)
        song_options_popup-add_to_playlist_button.tap()
        add_playlist_screen-playlists.match('Emil').tap()

  # Check if the end of the song_list has been reached
  bottom = songs_screen-song_list.scroll("down")
  if bottom:
    done = True
\end{lstlisting}











\subsubsection{Validation-based Script Revising.}
\label{sec:error_handling}
For each generated script, we validate it by executing it in a real environment before including it in the training dataset. 
Many of the generated scripts may not be executable due to LLM errors and app dynamics. Therefore, we need to handle encountered errors and revise the scripts accordingly during script validation. 
We introduce a domain-specific code status detector module that addresses errors by regenerating the script based on the error information. 
Specifically, for each task, we initially generate a code sample \( C = C_0 \), which is executed on a mobile device or emulator based on the domain-specific library mentioned in Section~\ref{sec:dsl}. 
If execution fails with error information \( \mathcal{E} \), it indicates issues with \( C \). The code status detector sends error information \( \mathcal{E} \) as well as the GUI state where the error occurs to the LLM, which is responsible for regenerating the script based on this GUI state to continue execution. 

Specifically, we categorize error \( \mathcal{E} \) as sending an illegal action to a GUI element (such as inputting text into a button), matching or indexing a non-existent element in the screen (such as index out of range for an element list), or logic errors (the previous action does not result in the GUI state that contains the current element). This categorization makes it easier for LLMs to understand the error, rather than processing raw error information.


\subsubsection{Tree-based Script Quality Improvement}
\label{sec:tree_based_search}

Even if a script executes successfully, it may not align with the intended user tasks. 
Thus, \name uses an on-cloud LLM-based reward model~\cite{judge_llm_as_a_judge} to evaluate whether the code \( C \) successfully achieves the task \( \mathcal{T} \). 
This reward model takes as input the task \( \mathcal{T} \), the executed code \( C \), and a sequence of environment observation states \( \mathcal{S} = (s_1, \dots, s_n) \), outputting a judgment \( \mathcal{J} \) and feedback \( \mathcal{F}_\text{out} \). 
If the reward model deems the code \( C \) as complete, the task-code pair \( (\mathcal{T}, C) \) is added to the dataset. Otherwise, the error information \( \mathcal{E} \) and feedback \( \mathcal{F}_\text{out} \) are recorded and used in a code search tree to regenerate a code sample. 

The code search tree is a Depth-First Search (DFS) tree designed to generate code samples, where each node represents a possible attempt to achieve the task goal through code. The initial code \( C_0 \) resides at the root node of the tree. During the search process, new nodes are expanded in a depth-first manner. At each node, the LLM is queried to generate a new code \( C' \) based on the task \( \mathcal{T} \), the current node’s code \( C \), the error information \( \mathcal{E} \), and the feedback \( \mathcal{F}_\text{out} \) from the reward model. The LLM generates new code sample \( C' \) that tries to address the error information \( \mathcal{E} \) and the feedback \( \mathcal{F}_\text{out} \), while achieving the task \( \mathcal{T} \).

The new code sample \( C' \) is then validated again through the dynamic task execution validator. The search continues until a code sample \( C \) is accepted by the validator or the maximum number of search attempts is reached. The final task-solution pair \( (\mathcal{T}, C) \) is then added to the final dataset, which will be used to fine-tune the SLM.

\subsection{Script-based Runtime Task Execution}
\label{sec:runtime}

At runtime, \name queries the LLM to synthesize domain-specific script tailored to the user's task, which can be executed by the domain-specific code executor. 

\subsubsection{Runtime Dynamicity Handling}

\name handles runtime dynamicity with two techniques. The first is the error handler, which is to re-generate the script after encountering runtime failures as described in Section~\ref{sec:error_handling}. The second is the dependency-aware instruction execution. 

At runtime, \name executes instructions from the script by grounding the target GUI element in the GUI screen to send proper GUI actions. 
\name introduces a dependency-enhanced instruction execution mechanism to handle incomplete scripts that may fail to account for all necessary navigation paths to a target GUI element. 
Specifically, let $e_c$ represent the GUI element invoked in the script, and let $T_c$ represent the current GUI tree, we match $e_c$ by checking: 1) Whether $T_c$ is the GUI state that $e_c$ belongs to, in case the current GUI screen is another state that contains different GUI elements that share the identical XPaths with $e_c$. 2) The identifier of $e_c$ can match one GUI element $e_i$ in $T_c$. 
If not matched, \name leverages \textit{backward dependency} of $e_c$ to automatically navigate to it. 
The dependency represents potential navigation sequences that can guide the GUI testing process to the desired element. 
Specifically, \name iteratively selects and executes actions from the dependency paths that are most likely to align with the current GUI context. 
If the element is still not found after completing all dependencies, \name raises an error and delegates the issue to the error handler module, signifying potential deficiencies in the app document or code logic.

\begin{algorithm}
\footnotesize
\caption{Dependency-Aware Element Locating}
\label{algo:dependency-aware-locator}
\begin{algorithmic}[1]
    \Require Target element $E$, current app instance $App$, dependency paths $\mathcal{D} = [D_1, D_2, \dots, D_k]$, app document $Doc$
    \Ensure Execution status $S$

    \Function{ElementGrounding}{$E, App, \mathcal{D}$}        
        
        \While{$E.identifier \notin current\_ui\_tree$ \textbf{and} \textit{attempts} $< \textit{limit}$}
            
            \State $Dependency \gets$  Select from $\mathcal{D}$  based on which dependency includes an action matching the current GUI state
            \For{action in $Dependency$}
                \State $current\_ui\_state \gets \text{GetCurrentUIState}(App, Doc)$
                
                \If{$action.element.ui\_state = current\_ui\_state$}
                    \State $App.send\_action(Action)$
                \EndIf
            \EndFor
        \EndWhile
        
        \If{$E \notin Current$}
            \State \Return Error \Comment{Section \ref{sec:error_handling}}
        \Else
            \State \Return $CurrentUITree.FindElement(E.identifier)$
        \EndIf
    \EndFunction
\end{algorithmic}
\end{algorithm}

\subsubsection{Enhancing Efficiency with Prompt Compression and Reuse}

Due to the limited computing resources of mobile devices and constrained context lengths of on-device SLMs, using the detailed app documents for task script generation (like how we synthesize data in Section~\ref{sec:dsl}) can be costly or even infeasible at runtime.
Therefore, we propose to shorten the runtime query prompt. Specifically, a runtime code query prompt consists only of the basic instruction, the user task, and all the element names in the app document, which reduces the prompt length from 15.4k to 2.8k on average (most of which can be cached and reused). 
The feasibility of such simplifications is from fine-tuning, which trained the intrinsic task-agnostic knowledge into the model parameters and left only the task-dependent in the prompt for the customized SLMs' reference for different tasks.
As such, this reduction does not affect the performance of on-device SLMs.

\name also employs prompt cache~\cite{llama_cpp} techniques to further speed up on-device SLM inference, particularly in reducing the prefilling latency. The prompt cache stores the KV state of existing queries, enabling a new query to reuse the KV cache if it shares the same prefix with a previous query, which allows the new query to skip the computation of the shared part. The document portion of a prompt in \name is consistent across all queries and can be cached and reused for each user task. Additionally, the document statement accounts for most of the input prompt length (by 97.6\%), so caching this part significantly reduces the prefilling latency. 

\section{Evaluation}
\label{sec:experiment}

We implement \name using Python and Java. SLMs are deployed on mobile devices using llama.cpp, a framework to enable LLM/SLM inference in C/C++. 
\wh{\name collects current GUI information via ADB (Android Debug Bridge). For each GUI action API, ADB commands are sent to the identified GUI element.}





\subsection{Experimental Setup}
\label{eval:setup}
\textbf{Dataset.}
We evaluate the effectiveness and efficiency of \name and baseline on two benchmarks: DroidTask~\cite{autodroid} and AitW-subset~\cite{AITW_dataset}. 
DroidTask~\cite{autodroid} is a mobile task automation dataset with 158 high-level tasks across 13 popular apps. 
AitW~\cite{llamatouch} is a large-scale dataset for Android device control. 
The original AitW dataset contains 417k tasks but lacks the exploration environment of the apps. 
So we choose the `google apps' and `general' subsets of the AitW dataset with 68 tasks using LlamaTouch~\cite{llamatouch} environment, which is executable and can be explored. The exploration trace is provided by the original datasets. 
Our document generation and task synthesis are independent of the test set.

\textbf{Training Details. }
\wh{We perform supervised fine-tuning on our model using the automatically generated task-solution training set detailed in Section~\ref{sec:tune_llm}, training for one epoch with a batch size of 32, a learning rate of 1e-5, and a maximum sequence length of 8192. For the baseline AutoDroid~\cite{autodroid}, we use the same training parameters.}

\textbf{Hardware.}
We evaluate the performance of \name on two devices: (1) A OnePlus ACE 2 Pro with 8 3.2 GHz ARM-based cores (Snapdragon 8 Gen2 CPU) and Adreno™ 740 GPU. 
(2) a MacBook Pro with an Apple M2 Pro chip, featuring a 10-core CPU, a 16-core GPU, and 16 GB of unified memory, selected to represent devices with higher computational capacity. 
The local LLM Llama-3.1 \cite{llama3} is deployed on the smartphone based on the llama.cpp framework \cite{llama_cpp}. 
The local LLMs are fine-tuned on an 8$\times$ A100 80GB server for 1 epoch, taking about 2.5 GPU hours.

\textbf{Baselines.}
We mainly compare \name with 2 types of baselines. 
(1) VLM-based agents, including CogAgent~\cite{cogagent}, SeeClick~\cite{cheng2024seeclick}. 
CogAgent is an 18B visual language model specializing in GUI navigation tasks. SeeClick is a GUI agent built on Qwen-VL~\cite{Qwen-VL}, with 9.6B parameters. 
(2) LLM-based agents, including AutoDroid~\cite{autodroid}, Mind2Web~\cite{mind2web}. 
AutoDroid is a memory-augmented mobile task automation framework, which is similar to our framework but adopts a step-wise method for task automation. 
We evaluate AutoDroid using GPT-4o~\cite{gpt4o} and a fine-tuned Llama-3.1~\cite{llama3} based on its automatically generated dataset.
Mind2Web is a generalist agent for UI task automation, which we evaluate based on GPT-4o~\cite{gpt4o} and Llama-3.1~\cite{llama3} respectively. 




\textbf{Metrics.}  
Given a sequence of UIs \(\{U_1, U_2, ..., U_n\}\) and corresponding actions \(\mathcal{A} = \{A_1, A_2, ..., A_n\}\) performed by human annotators to complete a task \(T\), if an agent generates a sequence of decisions \(\hat{\mathcal{A}} = \{\hat{A}_1, \hat{A}_2, ..., \hat{A}_k\}\) on \(\{\hat{U}_1, \hat{U}_2, ..., \hat{U}_k\}\), we use the following metrics to evaluate its performance:

\begin{itemize}[leftmargin=*]
\item \textbf{Success Rate:} In DroidTask~\cite{autodroid}, a task is considered completed if the ground-truth action sequence \(\mathcal{A}\) is a subsequence of the agent-generated sequence \(\hat{\mathcal{A}}\), i.e., \(\mathcal{A} \in \hat{\mathcal{A}}\). This metric reflects the agent's ability to complete the task. 
The LlamaTouch framework~\cite{llamatouch} evaluates task completion by sequentially comparing a task execution trace \(\{\hat{U}_1, \hat{U}_2, ..., \hat{U}_k\}\) against selected essential states \(\{U_{i_1}, U_{i_2}, ...\}\). A task is considered complete if the task execution trace includes all the essential states in order. 

\item \textbf{Reversed Redundancy Ratio:} As defined in AndroidArena \cite{xing2024understanding}, the reversed redundancy ratio is the ratio of the length of the ground-truth action sequence to the length of the agent-generated sequence, namely $n/k$. This metric measures the agent's efficiency when completing a task. 

\item \textbf{LLM Inference Latency: } We primarily measure the latency of LLM inference, which accounts for most of the system latency in on-device agents. Inference latency begins when the prompt is received by the LLM and ends when the final token of the output is generated. 

\item \textbf{Token Consumption: }We calculate token consumption based on the number of input and output tokens processed by the model. 

\end{itemize}




\subsection{Success Rate and Redundancy}
\label{sec:accuracy}
\subsubsection{DroidTask}

\begin{table*}
\caption{Success rate on DroidTask. \textbf{SR}: Success Rate. \textbf{RRR}: Reversed Redundancy Ratio. \textnormal{`ft' represents LLMs fine-tuned on GUI-specific dataset. }}
    \resizebox{1.0\textwidth}{!}
    {
        \begin{tabular}{cccccccccccccccc}
        \toprule
        \textbf{Metric} & \textbf{Method} & \textbf{Launcher} & \textbf{Calendar} & \textbf{Camera}   & \textbf{Clock} & \textbf{Contacts} & \textbf{Dialer} & \textbf{File} & \textbf{FireFox} & \textbf{Gallery} & \textbf{SMS} & \textbf{Music} & \textbf{Notes} & \textbf{Recorder} & Average\\
        \midrule 
        \multirow{5}{*}{SR}

        & SeeClick (Qwen-VL-9.6B-ft)  & 0\% & 0\% & 0\% & 9.1\% & 0\% & 0\% & 0\% & 0\% & 11.1\% & 13.3\% & 0\% & 0\% & 0\% & 2.7\% \\
        & CogAgent (CogVLM-17B-ft)    & 0\% & 0\% & 0\% & 9.1\% & 0\% & 0\% & 0\% & 0\% & 11.1\% & 13.3\% & 0\% & 0\% & 12.5\% & 3.3\% \\
        & Mind2Web (Llama-3-8B)   & 60.0\% & 41.2\% & 60.0\% & 50.0\% & 21.4\% & 33.3\% & 20.0\% & 12.5\% & 22.2\% & 40.0\% & 11.1\% & 35.7\% & 33.3\% & 34.4\% \\
        & AutoDroid (Llama-3-8B-ft)  &  80.0\% & 29.4\% & 50.0\% & 50.0\% & 21.4\% & 40.0\% & \textbf{43.8\%} & \textbf{87.5\%} & 44.4\% & \textbf{33.3\%} & 44.4\% & \textbf{35.7\%} & \textbf{66.7\%} & 43.9\%\\
        & \textbf{\name (Llama-3-8B-ft)}  & \textbf{80.0\%} & \textbf{64.7\%} & \textbf{66.7\%} & \textbf{83.3\%} & \textbf{50.0\%} & \textbf{60.0\%} & {31.2\%} & {75.0\%} & \textbf{66.7\%} & {26.7\%} & \textbf{55.6\%} & {28.6\%} & {55.6\%} & \textbf{54.4\%}\\
        
        \midrule


        

        \multirow{3}{*}{RRR}

        
        & Mind2Web (Llama-3-8B)  &  21.1\% & 26.7\% & 42.6\% & 29.6\% & 39.4\% & 43.0\% & 38.9\% & 66.7\% & 75.0\% & 58.3\% & 23.1\% & 42.7\% & 50.7\% & 41.0\%\\
        & AutoDroid (Llama-3-8B-ft)  &  73.8\% & 82.4\% & \textbf{92.3\%} & 70.8\% & \textbf{88.9\%} & \textbf{91.7\%} & \textbf{95.2\%} & 67.6\% & \textbf{100.0\%} & 95.0\% & 84.4\% & 95.0\% & 88.9\% & 86.3\%\\
        & \textbf{\name (Llama-3-8B-ft)}  & \textbf{93.8\%} & \textbf{94.7\%} & 91.7\% & \textbf{74.7\%} & 88.1\% & 88.9\% & 95.0\% & \textbf{76.4\%} & 96.7\% & \textbf{100.0\%} & \textbf{95.0\%} & \textbf{100.0\%} & \textbf{100.0\%} & \textbf{90.5\%}\\
        \bottomrule
        \end{tabular}
    }
    \label{tab:droidtask_acc}
\end{table*}

We first evaluate the accuracy of \name and baselines in DroidTask, and the result is shown in Table \ref{tab:droidtask_acc}. \name achieves an average task accuracy of 54.4\%, significantly higher than the baselines, whose accuracy ranges from 10.5\% to 43.9\%.
Among the baseline methods, AutoDroid~\cite{autodroid} achieves the highest performance because it also adopts automatic task generation and fine-tuning techniques to enhance the performance of on-device LLMs. VLM on-device UI agents (CogAgent~\cite{cogagent} and SeeClick~\cite{cheng2024seeclick}) exhibit lower accuracy because they rely solely on screenshot images, which are inadequate for text-heavy apps. Besides, the grounding capabilities of smaller on-device VLMs are limited, leading to frequent misplacement of UI elements.

\name outperforms baselines mainly because of the following reasons: 
Compared to methods that do not fine-tune LLMs, \name generates fine-tuned data that covers every GUI element in the app document, offering detailed insights into how to use the app. 
While AutoDroid~\cite{autodroid} also fine-tunes LLMs by generating tasks and solutions, it relies on a step-wise method. 
In contrast, \name uses script-based fine-tuning data, which is more suitable for smaller LLMs. 
For the script-based method, the responsibility of determining whether a task is complete is shifted from the LLMs to the code executor. If a script runs without errors, it is considered complete, thus eliminating the need for the LLM to make a task completion decision. 
And AutoDroid is less effective at determining task completion~\cite{autodroid}. AutoDroid tries to address this limitation by incorporating a small amount of manually annotated data from~\cite{motif}, this solution is neither scalable nor fully effective. Our experiments found that AutoDroid repeatedly executes actions for 25.5\% 
The success rates for the Files, SMS, and Notes apps in \name are lower than expected, primarily because some high-quality fine-tuning data for these apps was mistakenly filtered out by the solution validation module, a problem discussed in Section~\ref{sec:solution_validator}. In practice, this issue could be addressed by exploring the app and pre-setting some user data within it.

For the reversed redundancy ratio, systems based on fine-tuned LLMs (AutoDroid~\cite{autodroid}, \name) are better than systems based on original LLMs (Mind2Web~\cite{mind2web}). 
LLMs can get a comprehensive overview of apps after fine-tuning and can efficiently navigate to the specific UI elements that are most likely to solve the task. 
Seeclick~\cite{cheng2024seeclick} and Cogagent~\cite{cogagent} struggle to complete tasks in DroidTask, with their success rates being exceptionally low. Due to their inability to successfully complete most tasks, we do not include these two methods in the RRR statistics. 

\subsubsection{AitW}

\name outperforms the baselines by an average of 14.3\%, as shown in Table~\ref{table:aitw}. LLM-based GUI agents (CogAgent~\cite{cogagent} and SeeClick~\cite{cheng2024seeclick}) perform better on AitW~\cite{AITW_dataset} than on DroidTask~\cite{autodroid} because they were fine-tuned on the dataset and have a comprehensive understanding of the apps.
The results demonstrate that \name consistently outperforms the baselines in complex apps like Chrome, Google Maps, and YouTube. This is because these apps often feature dynamic or deeply nested UI elements (\eg hierarchical menus in Google Maps or dynamic content in YouTube driven by recommendation systems). Step-by-step agents may struggle to locate and interact with such elements consistently due to the limited context provided by a single screen. In contrast, \name's script-based approach, with access to a full application overview, allows it to consider the potential effects of each action and anticipate future elements, leading to more effective solutions.

\begin{table}
\caption{Success rate on AitW-subset. \textbf{SR}: Success Rate.}
    \resizebox{.3\textwidth}{!}
    {
        \begin{tabular}{cccccccccc}
        \toprule
        \textbf{Method} & \textbf{Model} & \textbf{SR}   \\
        \midrule 
        Mind2Web  &Llama-3.1-8B           & 27.1\%   \\
        
        SeeClick  &Qwen-VL-9.6B           & 30.8\%  \\
        CogAgent  &CogGLM-17B-ft         & 36.7\%  \\
        AutoDroid &Llama-3.1-8B-ft        & 36.7\%  \\
        \textbf{\name}     &\textbf{Llama-3.1-8B-ft}        & \textbf{47.1\%}  \\
        \bottomrule
        \end{tabular}
        
    }
\label{table:aitw}
\end{table}

\subsection{Latency and Cost}

\begin{figure}

    \centering
    \begin{subfigure}[b]{0.5\textwidth}
        \centering
        \includegraphics[width=\textwidth]{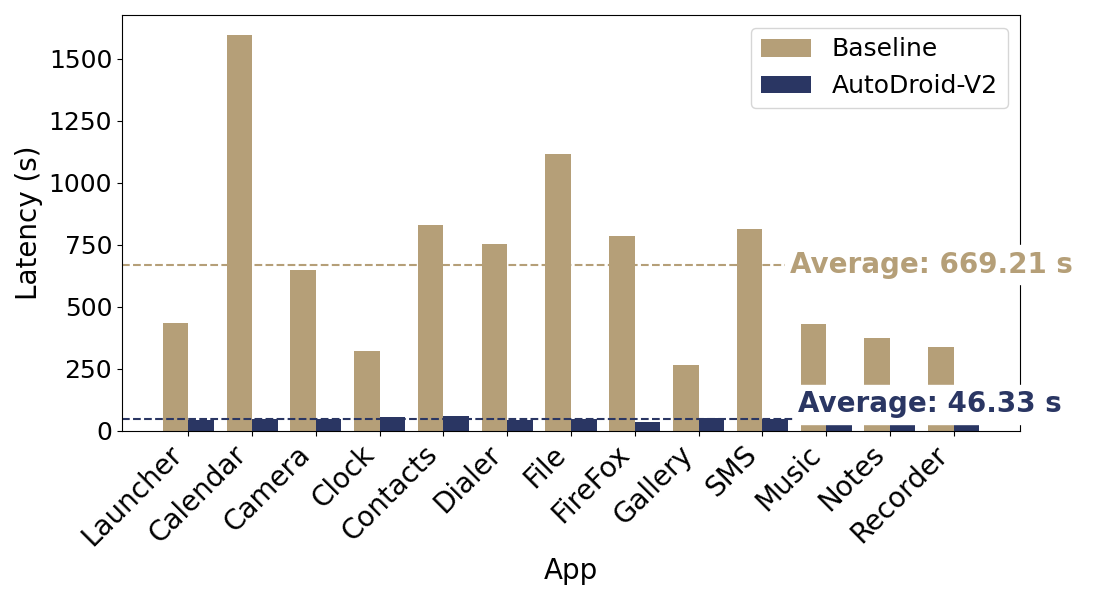}
        \caption{LLM inference latency in Snapdragon 8 Gen 2.}
    \end{subfigure}
    \hfill
    \begin{subfigure}[b]{0.5 \textwidth}
        \centering
        \includegraphics[width=\textwidth]{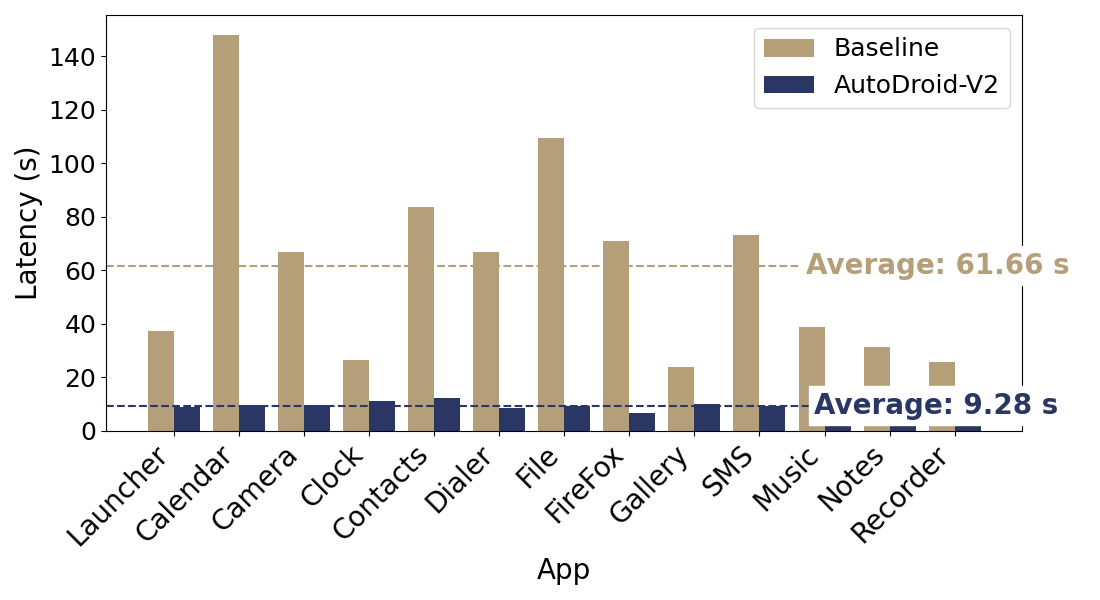}
        \caption{LLM inference latency in Apple M2 Pro.}
    \end{subfigure}
    \caption{Average LLM inference latency of \name and baseline (AutoDroid) on DroidTask.}
    \label{fig:latencies}
\end{figure}

Figure~\ref{fig:latencies} illustrates the on-device LLM inference latencies of \name and the step-wise method (AutoDroid~\cite{autodroid}) across different apps on Snapdragon 8 Gen 2 and Apple M2 Pro. We quantize the fine-tuned Llama3.1 8B models of \name and AutoDroid to 8-bit precision and deploy them on mobile devices using the Llama.cpp framework~\cite{llama_cpp}.
The average LLM inference latency for each task of \name is 46.3s, compared to 669.2s for the baseline on Snapdragon 8 Gen 2, resulting in a 93.1\% reduction in inference latency. 

This acceleration is mainly due to the script-based method, which reduces the number of LLM queries. 
For the step-wise method, the latency for each step remains stable, so the overall latency depends on the number of steps in a task. The Calendar app, for instance, has the highest average step count (13.4 steps per task), resulting in the highest latency. As the number of steps increases with more complex tasks, the latency for step-wise methods rises much faster than for the script-based method.

Table~\ref{tab:runtime-token-consump} shows the average runtime token consumption for \name and the baseline method. The cached tokens are the common prefix part that has already been processed, and the internal state of LLM has been stored in a file, which can be pre-loaded to memory and reused. 
The cached tokens refer to the common prefix that has already been processed, with the LLM's internal state stored in a file, allowing it to be preloaded into memory and reused. The remaining tokens are part of the input prompt following the cached prefix, which must be processed in real-time during model inference. 
For the step-wise method, the main part of a prompt is the GUI state, which cannot be cached because they are dynamic and frequent changes. 
Conversely, \name primarily uses the app document, which is static and can be cached. 
The latency for each response mainly depends on the remaining and output tokens. \name saves 97.8\% of the remaining tokens and 85.2\% of the output tokens, significantly reducing the computational burden.

The large portion of the common prefix in \name prompts allows for the use of the prefix key-value (KV) cache, which greatly reduces prefill time. LLM inference latency arises primarily from two phases: the prefill phase, where input text is prepared, and the decoding phase, where the model generates output based on the input and prior tokens.
The prefill latency for \name is significantly reduced compared to the baseline (14.0s versus 85.3s per prompt) because most of the prompt has already been cached.

We examine token consumption and associated costs of \name while utilizing GPT-4o\cite{gpt4o} across three stages: document generation, data synthesis, and solution validation. These stages represent a one-time data collection process, during which GPT-4o is used to prepare the necessary data. Once the data are collected, the task execution process of \name does not require the intervention of GPT-4o. In the document generation stage, the average consumption of input tokens per application is 1.28M, and the average consumption of output tokens per application is 91.0K, resulting in a cost of \$4.11 per application. The data synthesis stage consumes 2.98M input tokens and 38.2K output tokens per application at a cost of \$7.83 per application. The solution validation stage demands the most substantial computational resources and incurs the highest costs, with an average consumption of 26.81M input tokens and 343.9K output tokens per application, leading to a cost of \$70.48 per application.

\begin{table}
\caption{Runtime token number of \name and AutoDroid. \textnormal{`Cached' denotes the cached prompt prefix that has been processed and can be reused. `Remaining' denotes the remaining parts of the input prompt that come after the cached prefix. }}
    \resizebox{.45\textwidth}{!}
    {
        \begin{tabular}{cccccccccc}
        \toprule
        & \multirow{2}{*}{\textbf{Method}} & \multicolumn{3}{c}{\textbf{Input Tokens}} & \multirow{2}{*}{\textbf{Output Tokens}} \\
        \cmidrule(l){3-5}
        &                 & \textbf{Cached} & \textbf{Remaining} & \textbf{Total}  & \\
        \midrule 
        \multirow{2}{*}{Per Step}
        & AutoDroid        &  66.0  &  452.1 &  518.1 &  127.4    \\
        \cmidrule(lr){2-6}
        & \name            &  N/A &  N/A &  N/A &  N/A  \\
        \midrule
        \multirow{2}{*}{Per Task}
        & AutoDroid           &  431.3 &  3021.2 &  3452.5 &  832.4 \\
        \cmidrule(lr){2-6}
        & \name               &  2760.1 &  \textbf{67.9} & 2828.0  &  \textbf{122.9} \\
        \bottomrule
        \end{tabular}
        
    }
    \label{tab:runtime-token-consump}
\end{table}

        

\subsection{Fine-grained Performance Analysis}

\subsubsection{Performance with Different LLMs}
\begin{figure}
  \centering
  \includegraphics[width=0.95\linewidth]{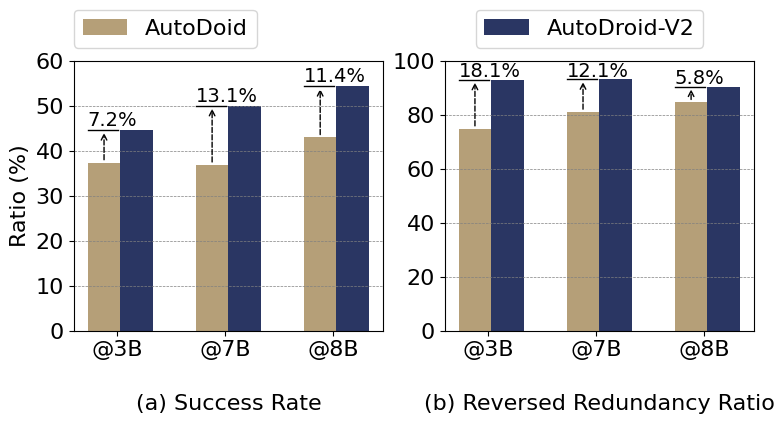}
  \caption{Success rate and reversed redundancy ratio of \name and baseline based on different LLMs. llama3.2-3b, qwen2.5-7b, llama3.1-8b}
  \label{fig:sr-rrr-diff-LLM}
\end{figure}

We use AutoDroid, the best-performing baseline on the Llama-3.8B model, as a reference and compare it with \name across different LLMs.
The results presented in Figure~\ref{fig:sr-rrr-diff-LLM} show the comparison of success rate and reversed redundancy ratio between \name and the baseline models for different LLMs: Llama3.2-3B, Qwen2.5-7B, and Llama3.1-8B, where \name achieves 44.6\%, 50.0\%, 54.4\% success rate respectively. 
The reversed redundancy ratio of \name is stable, ranging from 90.5\% to 93.0\%. This is due to the generated scripts being designed to be straightforward and focused on task completion without including exploration or trial-and-error steps. 

\begin{table}
\caption{Success rate on DroidTask based on on-cloud LLM (GPT-4o). }
    \resizebox{.45\textwidth}{!}
    {
        \begin{tabular}{cccccccccc}
        \toprule
        \textbf{Dataset} & \textbf{Metric} & \textbf{Mind2Web} & \textbf{AutoDroid} & \textbf{\name}  \\
        \midrule 
        \multirow{2}{*}{DroidTask}
        & SR        &          65.2\%   &         62.7\%    &  67.1\%   \\
        & RRR   &  92.5\%   &         92.3\%    &  90.1\%    \\
        \midrule
        \multirow{1}{*}{AitW-subset}
        & SR        &          41.1\%   &         30.9\%    &  51.4\%   \\
        \bottomrule
        \end{tabular}
        
    }
    \label{tab:oncloud_sr}
\end{table}



We also conduct experiments using powerful on-cloud LLMs (GPT-4o) as a reference for comparison with on-device LLMs, as shown in Table~\ref{tab:oncloud_sr}. The method employed by \name is described in Section~\ref{sec:tune_llm}. The success rate of \name is 9.3\% higher than the baselines on average across two benchmarks. We summarize the failure cases observed with the \name below:

\textbf{Match Errors}: The <element>.match(<text or attribute dict>) syntax finds target elements in the list that match the provided text or attribute dictionary. However, the provided texts may not always align with the actual element text. For example, in the task \textit{``change the theme of the app to black''}, the LLM might generate \textit{theme\_options.match('black').tap()}, but the available options could be \textit{``white'', ``dark'', or ``auto light/dark''}. In this case, the correct action would be to tap ``dark''. A solution is to use a lightweight embedding model to find the most similar text to the input. 

\textbf{Misunderstanding Syntax}: We have observed instances where certain code statements are generated incorrectly. For example, \textit{favorite\_button.match(`selected') == `false'} should be written as \textit{favorite\_button.match(`selected') == False}. Enhancements to the code generation capabilities of the foundation model could help address such issues.

\textbf{Neglecting Confirmation Elements}: Some tasks require confirmation at the final step, such as tapping elements like "save" or "confirm." These confirmation actions are sometimes overlooked in the generated code. 

\subsubsection{Impact of Script Validation}
\label{sec:solution_validator}


We also compare the success rate of \name with and without the train data validator module mentioned in Section~\ref{sec:error_handling} and Section~\ref{sec:tree_based_search}. 
The LLM is Llama3.1 8B, and the average success rate of methods with and without a solution validator are 54.4\%, and 51.9\% respectively, improved by 2.5\%. 
We observe that 4 applications such as Contacts, Files, Notes, and SMS experience a minor decrease in success number (1-3 tasks) after the validation step. These apps are highly user-specific, containing personal data such as contact names, file names, and notes. Validation becomes problematic for these apps because some tasks cannot be completed unless certain user-specific data is present. For instance, the task "Delete a contact named Alice" cannot be completed if Alice is not in the user's contact list. 
Thus, some valid tasks and solutions are filtered out by the validator, even though the solutions themselves are correct. Consequently, these tasks are excluded from the training data, which diminishes the fine-tuned LLM's ability to complete such tasks. 
For other apps, tasks do not require specific user data (\eg Camera, Calendar, Recorder). In these cases, the validator can filter out incorrect solutions, improving or maintaining the fine-tuned model's accuracy. 
One potential solution to this issue is to pre-set user data using a random explorer before validating the app. The task generator can then be provided with this user data, enabling it to generate tasks that can be validated successfully.

\section{Discussion}
One potential concern is how \name handles situations where structured text representations of GUIs are unavailable. Typically, \name identifies and interacts with the target GUI elements at runtime by matching their identifiers, which are text-based XPaths. However, some special mobile apps, such as Unity-based apps and Web-based apps, may not provide structured text representations. 
A possible solution is to recover structured GUI representations based on the visual features.
Vision-Language Models (VLMs) have shown their great potential in this field, and it is foreseeable that structured GUI representations can be obtained by VLMs accurately and efficiently.

Another possible limitation of \name is the difficulty of handling highly dynamic user interfaces, such as those in web search engines and games. In these apps, it is difficult to plan the actions in advance, since the future states are usually unpredictable. 
In concept, \name can deal with such dynamic user interfaces, but it may need to regenerate the plan for each step, leading to degraded efficiency.
Our future work will focus on how to seamlessly integrate the step-wise reasoning approach with our script-based approach. Future agents are expected to flexibly switch between the two modes for different types of user interfaces, handling dynamic content effectively while maintaining efficiency.
\section{Conclusion}

We present a document-guided, script-based, end-to-end system named \name to support mobile task automation using on-device SLMs. Experiments show that the script-based method significantly improves the efficiency and performance of GUI agents. We believe this approach has the potential to enable the full deployment of GUI agents on devices, achieving accuracy comparable to that of cloud-based GUI agents.

\section*{Acknowledgement}
\label{sec:acknowledgement}
This work is supported by the National Key R\&D Program of China (No.2022YFF0604500), National Natural Science Foundation of China (Grant No.62272261), Xiaomi Foundation, Wuxi Research Institute of Applied Technologies, Tsinghua University (Grant No.20242001120), and a grant from the Guoqiang Institute, Tsinghua University. 

\balance
\bibliographystyle{ACM-Reference-Format}
\bibliography{reference}

\end{document}